\documentclass{article}
\usepackage{spconf,amsmath,graphicx}
\usepackage{bm}
\usepackage{multirow}
\usepackage{booktabs}
\usepackage{enumitem}
\usepackage{paralist}
\usepackage{xcolor}
\graphicspath{ {figure/} }
\newif\ifsubmit

\submitfalse
\ifsubmit
\newcommand{\sychien}[1]{}

\newcommand{\ctliu}[1]{}

\else
\newcommand{\sychien}[1]{{\bf \textcolor{purple}{S.-Y. Chien: #1}}}

\newcommand{\ctliu}[1]{{\bf \textcolor{brown}{C.-T. Liu: #1}}}

\fi

\newcommand{\etal}{\textit{et al}.}
\title{Interactive Object Segmentation with Dynamic Click Transform}
\name{Chun-Tse Lin, Wei-Chih Tu, Chih-Ting Liu, Shao-Yi Chien}
\address{Graduate Institute of Electronics Engineering, National Taiwan University, Taipei, Taiwan}
\begin{document}
%\ninept
%
\maketitle
\begin{abstract}
In the interactive segmentation, users initially click on the target object to segment the main body and then provide corrections on mislabeled regions to iteratively refine the segmentation masks.
Most existing methods transform these user-provided clicks into interaction maps and concatenate them with image as the input tensor.
Typically, the interaction maps are determined by measuring the distance of each pixel to the clicked points, ignoring the relation between clicks and mislabeled regions.
We propose a Dynamic Click Transform Network~(DCT-Net), consisting of Spatial-DCT and Feature-DCT, to better represent user interactions.
Spatial-DCT transforms each user-provided click with individual diffusion distance according to the target scale, and Feature-DCT normalizes the extracted feature map to a specific distribution predicted from the clicked points. 
We demonstrate the effectiveness of our proposed method and achieve favorable performance compared to the state-of-the-art on three standard benchmark datasets.
\end{abstract}

\vspace{-4mm}
\section{Introduction}
\vspace{-3mm}
\label{sec:intro}

Interactive segmentation, also known as interactive object selection, aims to segment
the object of interest and refines the segmentation mask via humans-in-the-loop.
The segmentation results are useful for many applications, such as video editing, medical image analysis, and especially human-machine collaborative annotation.
Because the demand for fine-grained image annotations dramatically increases with the development of data-driven deep learning methods, an efficient interactive segmentation is in need to alleviate the burden of manually labeling each pixel in an image.
%
%However, manually labeling each pixel in the image is a laborious process, and thus, the efficient interactive segmentation algorithms are in need to alleviate the annotating burden.
%

Among the interactive segmentation scenarios, user interactions are usually given through bounding boxes~\cite{rother2004grabcut},  clicks~\cite{xu2016deep,jang2019interactive,li2018interactive,liew2017regional}, or scribbles~\cite{grady2006random,gulshan2010geodesic}.
A box-interfaced one lets the user indicate the target by drawing a bounding box to obtain the entire object's information. 
However, some background pixels are included at the same time, making the user intention imprecise. 
In contrast, a click-interfaced one gets a precise location but lacks object/region size.
Although scribbles can get precise and rich information, drawing scribbles places much more burden than clicking points. 
%
%In our paper, we propose a new interaction based on the click interface which can contain more information without much burden and at the same time be consistent and easy for users.
%In our paper, we propose a consistent and easy interaction based on click interface that gets more information without much burden.
% \ctliu{Need a simple summary here. Because it seems that click is easy but you say with a lack of size.}
%
Given those user inputs, classical approaches~\cite{boykov2001interactive,rother2004grabcut,bai2007geodesic,grady2006random} formulate the segmentation process as a graph-based optimization problem. %, taking each pixel as a vertex on the graph and compute the affinity between pixels as the edges.
%
% Boykov and Jolly~\cite{boykov2001interactive} formulate interactive segmentation as graph cut optimization, and Rother et al.~\cite{rother2004grabcut} extend graph cut by iterative fitting foreground and background distribution.
% %
% Bai and Sapiro~\cite{bai2007geodesic} determine foreground and background by computing the weighted geodesic distance between unlabeled pixels and user-provided scribbles.
% %
% Grady~\cite{grady2006random} proposes a method that starts a random walker at each pixel and assigns it to the label of the first seed that the walker reaches.
%
%ctliu
%These algorithms rely on low-level image features and hand-crafted distance metrics, which may not be effective to distinguish the target object from the whole image. 
%
%ctliu
%Thus, it results in a large number of interactive inputs from a user to obtain a reliable result.
%Thus, it requires a large amount of input from a user to obtain a reliable segmentation, increasing the user’s burden.
%
%
% Recently, deep neural networks have shown superior performance in many computer vision tasks, and the deep feature learned from DNNs are highly transferable between different problems. 
%
Inspired by the success of fully convolutional neural networks (FCNs)~\cite{long2015fully} on semantic segmentation, Xu~\etal~\cite{xu2016deep} first proposed a deep learning-based interactive segmentation algorithm.
They compute two additional distance maps representing positive and negative clicks from the user and concatenate them with the input image to generate the desired foreground mask with an FCN model.
%ctliu
%Then, the desired foreground mask is predicted by an FCN model trained on these image-interaction pairs.
%Then the probability that each pixel belongs to the foreground is predicted by an FCN model trained on these image-interaction pairs.
%
% \ctliu{Too many Related work, you can choose some important works and describe more about the weakness}
Most later works follow this strategy but transform the user clicks into Euclidean distance maps~\cite{xu2016deep, liew2017regional}, Gaussian~\cite{li2018interactive, mahadevan2018iteratively}, or multiple guidance maps~\cite{majumder2019content}, respectively.
%
% Liew~\etal~\cite{liew2017regional} proposed a RIS-Net to capture regional information from patches that include pairs of positive and negative clicks for refining global prediction.
%
% Li et al.~\cite{li2018interactive} produce multiple hypothesis segmentation masks and select one of them to solve ambiguity.
%
% Maninis et al.~\cite{maninis2018deep} introduce a novel interactive way that requires user clicks on extreme points.
%
% Mahadevan et al.~\cite{mahadevan2018iteratively} propose an interactively training scheme to increase the correlation between clicks and prediction errors. 
% %
% Majummder and Yao~\cite{mahadevan2018iteratively} transform user clicks into multiple guidance maps as input to neural networks according to other information, such as superpixel.
%
To further make use of user clicks, BRS~\cite{jang2019interactive} and f-BRS~\cite{sofiiuk2020f} proposed a back-propagating refinement scheme to adjust the original input clicks maps by forcing the interaction points to have the correct predicted labels. % by adjusting the input clicks maps.
However, these methods regard all clicks as equal importance and transform them with an identical function, discarding the relation between clicks and the target object.
Moreover, the back-propagating-based methods need to additionally minimize the predefined energy function through backpropagation iteratively, which includes extra computation and increases the inference time.
%The optimization process in these works needs to minimize the predefined energy function through backpropagation iteratively, which includes extra computation and increases the inference time.

\begin{figure*}[htb]
\includegraphics[width=1.0\linewidth]{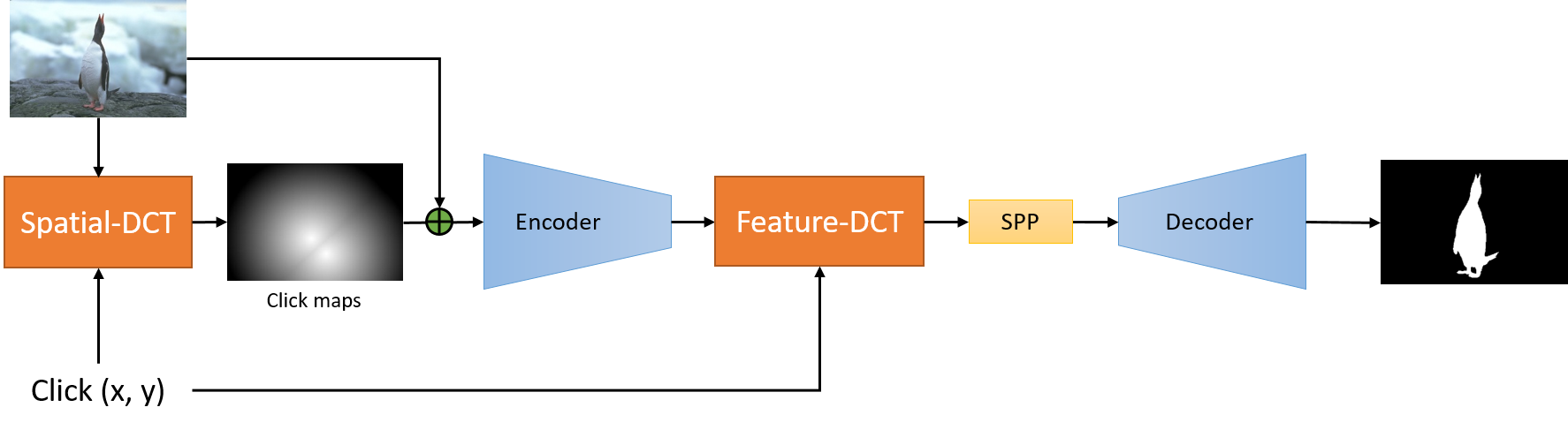}
\caption{Overview of Dynamic Click Transform Network. Spatial-DCT dynamically encodes user interactions into distance map and Feature-DCT scales and shift the original feature for better prediction.}
\label{fig:arch}
\end{figure*}
%In our paper, we propose a new interaction based on the click interface which can contain more information without much burden and at the same time be consistent and easy for users.
In this paper, we start from adopting a click-based interaction firstly proposed in \cite{pont2015semi}, called \textit{Click-and-Drag}. It adds a drag action for each click, which is nearly without extra burden for users. 
This novel interaction scheme combines the advantage of click and bounding box to get precise location and contain more object scale information.
Then, we propose a Dynamic Click Transform Network (DCT-Net), which contains two components, a Spatial Dynamic Click Transform~(Spatial-DCT) and a Feature Dynamic Click Transform~(Feature-DCT).
This network takes both spatial geometry and feature distribution into consideration to make good use of click-and-drag interaction.
%
%DCT-Net consists of a Spatial Dynamic Click Transform~(Spatial-DCT) and a Feature Dynamic Click Transform~(Feature-DCT).
%
Spatial-DCT transforms each user click into 2D maps by applying an individual Gaussian mask which is dynamically determined by the object scale.
Compared to the identical transform used in most previous works, our approach is more robust to object in different scales.
Feature-DCT further uses the user clicks in the feature domain by refining the whole feature distribution of the input image according to the feature at the clicked position.
With this operation, the feature changes dynamically in each interaction, helping focus on some mislabeled parts.
The main contributions of this paper are:
\begin{compactitem}
\item We adopt a Click-and-Drag interaction, which can take advantage of both click and bounding box.

\item We propose a Spatial Dynamic Click Transform to encode both the object scale and refine region into the distance maps.% and provide two ways to obtain the diffusion size.
\item We propose a Feature Dynamic Click Transform to aggregate all clicked features and adjust the whole image feature to distinguish pixels belonging to the object of interest.
\end{compactitem}

\vspace{-2mm}
\section{Proposed Method}
\vspace{-2mm}
\label{sec:method}
The architecture overview is illustrated in Fig.~\ref{fig:arch}.
The proposed Dynamic Click Transform Network~(DCT-Net) is based on an encoder-decoder architecture with spatial pyramid pooling (SPP).
%We propose the Dynamic Click Transform Network~(DCT-Net) for interactive segmentation.
%
%This network is based on an encoder-decoder architecture with spatial pyramid pooling.
%
We transform the user clicks not only in the spatial domain but also in the high dimensional feature domain by using the proposed Spatial-DCT and Feature-DCT, respectively.
In the Spatial-DCT, we encode each click by Gaussian mask with individual diffusion radius, determined from the target region size.
The Feature-DCT is performed by scaling and shifting the feature extracted from the input image, resulting in a better distribution for separating the target object.

\subsection{Spatial Dynamic Click Transform}
\label{ssec:dst}
\begin{figure}
\centering
\includegraphics[width=1.0\linewidth]{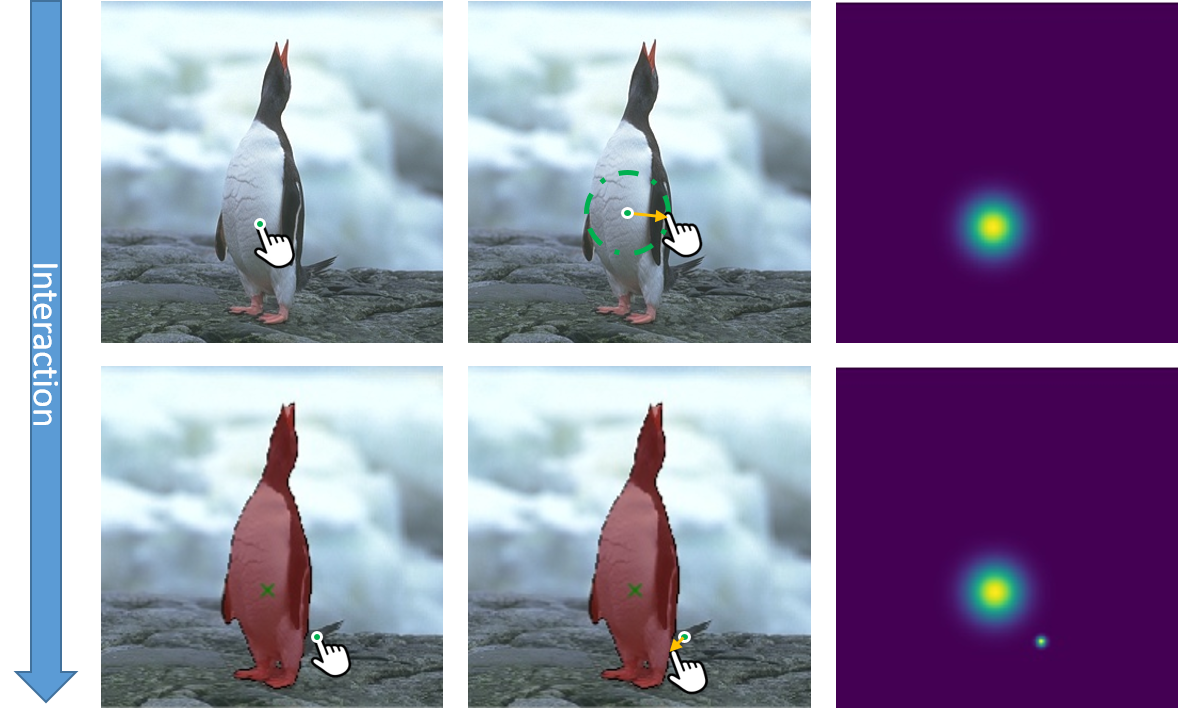}
\caption{Click and Drag scheme.}
\label{fig:c&d}
\end{figure}

Most previous works, which use either Euclidean distance transform or Gaussian transform, regard all clicks as equal importance. % and transform with an identical mask.
Given a sequence of user interactions $\mathcal{C}$ includes a positive click set $\mathcal{C}^1$ and a negative click set $\mathcal{C}^0$.
The clicks are encoded into two distance maps $\mathcal{D}^1$ and $\mathcal{D}^0$ for positive and negative clicks, respectively.
More formally, the pixel value at the location $\bm{p}$ can be computed as:
\begin{align}
    \mathcal{D}^{l}_{d}(\bm{p}) = \min_{\bm{c} \in \mathcal{C}^l} d(\bm{p},\bm{c}), l \in \{0,1\}
\end{align}
where $d(\cdot,\cdot)$ is the Euclidean distance or Gaussian map.
These distance maps only localize the user clicks and ignore the target object scale or mislabeled region size, which can directly impact network performance.
Instead, we take the relation between click and target object or mislabeled region into consideration and encode this information into the distance maps.
The function can be written as:
\begin{align}
    \mathcal{D}^{l}_{f}(\bm{p}) = \min_{\bm{c} \in \mathcal{C}^l} f(\bm{p},\bm{c}, r(\bm{c})), l \in \{0,1\}
\end{align}
We take a Gaussian function $f$ as our transform function and dynamically change the diffusion distance by an extra variable $r$ according to the user clicks.
% We take a Gaussian function $f$ as our transform function and take one more variable $r$, which is dynamically determined by the user clicks, as an extra input.
%
% The variable $r$ presents the click's diffusion distance, and re-weight the distance between the clicked point and other pixels.
%
The simplest and precise way to get a proper value of $r$ is obtaining from user interactions.
% %
% When the users click on the mislabeled region, they have already known the size of this region.
% %
% Thus, we can get this information without adding much burden and additional computation.
%
Illustrated in Fig.~\ref{fig:c&d}, we adopt a novel interaction interface called Click-and-Drag. 
The user is asked to click on the center of the largest incorrect region and then drag outward until reaching the nearest boundary. 
The distance between the click and released position is recorded as the diffusion distance of this click.
With the user-given drag, we can clearly understand the user's intention and know the size of the mislabeled region. %helping solve the ambiguity between clicks and the target object.
In our experiments, to fairly compare to other click-based methods, we also propose an Auto-Drag-Head, a lightweight neural network that can automatically predict the diffusion distance $r$.
Even with this predicted value, we can also perform a better result than that of using an identical transform.
% Under the standard click interaction, the diffusion distance is regressed from the current input click and previously predicted segmentation mask.

\vspace{-2mm}
\subsection{Feature Dynamic Click Transform}
\label{ssec:dft}
\begin{figure}[htb]
\includegraphics[width=1.0\linewidth]{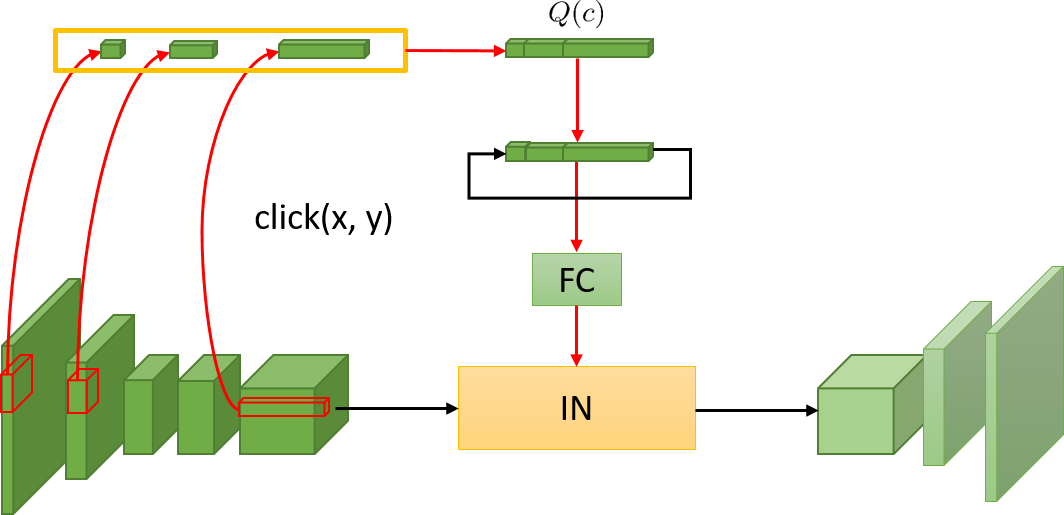}
\caption{Feature Dynamic Click Transform.}
\label{fig:s_dct}
\end{figure}

From the user input clicks, we can gather more information in addition to the spatial correlation.
In the Feature-DCT, we utilize the features extracted from the input image at the user clicked positions, which are rarely used in most existing methods.
Firstly, we gather the feature at the clicked position and feed it into a fully connected network to output a set of means and variances for each channel.
Secondly, the original feature map is scaled and shifted by the predicted means and variances, then fed into the segmentation head.
%
% Different from AdaptIS~\cite{sofiiuk2019adaptis}, which takes only the center point of the object to perform instance segmentation, 
When a new click comes, we apply a feature aggregation strategy to take all click points into account.
The aggregation is doing by vector sum if a positive click is given; otherwise, vector rejection is applied for a negative click.
Given a user correction clicked at $\bm{c}$, the aggregated feature $f$ can define as:
\begin{align}
\begin{cases}
    f_i = \frac{f_{i-1} + Q(\bm{c})}{2} & if ~~\bm{c} \in \mathcal{C}^1, \\
    f_i = f_{i-1} - (f_{i-1} \cdot Q(\bm{c})) \frac{Q(\bm{c})}{\|Q(\bm{c})\|} & if ~~\bm{c} \in \mathcal{C}^0
\end{cases}
\end{align}
where $f_i$ is the aggregated feature in the $i^{th}$ interaction, and $Q$ is the feature extracted from the input image.
Fig.~\ref{fig:s_dct} illustrates the Feature-DCT for the input image feature.
More detailed, we extract the feature in three different layers corresponding to the click position and concatenate these features.
And then aggregate this multi-level feature by the strategy mentioned above.
Last, the fully connected network predicts three sets of means and variance for applying instance normalization~(IN)~\cite{ulyanov2016instance} on the features of a U-net.
The correction level that each click focuses on and the difference between clicks at each interaction is efficiently used.
The Feature-DCT, thus, refines not only the internal region but the area near boundaries.

\subsection{Interactively Training}
\label{ssec:train}
For the Dynamic Click Transform Network to learn the relation between user correction and predicted segmentation, we train our network click by click, similar to that in~\cite{forte2020getting}.
Starting from a single click on the farthest pixel from the object boundary, a sequence of interactions is given according to the output mask.
%
% As the sequential scheme used to evaluate interactive segmentation algorithm, we place a click at the center of the largest incorrect region so as to maximize the Euclidean distance to the region boundary.
%
The loss is computed, and the weights are updated at each interaction.
Since user annotations are impractical to obtain from humans during training, we turn to simulate from the ground truth segmentation mask and network predicted mask.
For the first click, we compute the minimum distance to the object boundary for each pixel on the target object.
Then pick the farthest point from the boundary and take the corresponding distance computed above as the diffusion distance for spatial-DCT.
After the initial segmentation mask is predicted, we generate the subsequent clicks with respect to the previous prediction of the network.
%
% First, we identify the mislabelled pixels by comparing the output mask with the ground truth mask and group them into connected components.
%
A click is then sampled on the largest mislabelled region such that the euclidean distance from the boundary is larger than other pixels within this region.
Then the sampled click is considered a positive click if the corresponding pixel lies on the object or a negative click otherwise.

\section{Experiments}
\label{sec:exp}

\subsection{Experimental Settings}
\label{ssec:settings}
We evaluate our proposed method on three publicly available datasets: GrabCut~\cite{rother2004grabcut}, Berkeley~\cite{mcguinness2010berkeley} and DAVIS~\cite{perazzi2016davis}.
GrabCut contains 50 images and provides a single object mask for each image; pixels in a thin band around the object boundary are not valid.
Berkeley consists of 100 object masks on 96 images and represents some challenges encountered in interactive segmentation.
DAVIS contains 50 videos with high-quality ground truth masks.
To evaluate interactive segmentation algorithms, we use the same 354 individual frames sampled from videos as~\cite{jang2019interactive}.
%

% \subsection{Evaluation Metrics}
% \label{ssec:eval}
As for the evaluation, we use the same click generation strategy as in previous works and take a robot to simulate user clicks.
After each interaction, we calculate the intersection of union~(IoU) between the predicted mask and ground truth mask and plot the mean intersection of union~(mIoU) score according to the number of clicks.
Then, we adopt the mean number of clicks~(mNoC), which counts the average number of clicks required to achieve a target IoU threshold.
We set the IoU threshold as 90\%, and the default maximum number of clicks is limited to 20 for each sample, consistent with the previous works.

\begin{table*}
\caption{Ablation studies of proposed methods.}
\vspace{5pt}
\label{tab:ablation_arc}
\centering
\footnotesize 
\begin{tabular}{l|c|cc|cc|cc}
\toprule
\multirow{2}{*}{Method} & \multirow{2}{*}{Interaction} & \multicolumn{2}{c|}{GrabCut} & \multicolumn{2}{c|}{Berkeley} &\multicolumn{2}{c}{DAVIS} \\
\cline{3-8}
&& NoC @ 90\% & AuC & NoC @ 90\% & AuC & NoC @ 90\% & AuC \\
\hline\hline
Baseline & \multirow{3}{*}{Click and Drag} & 4.4 & 0.904 & 5.73 & 0.901 & 9.13 & 0.821 \\
Baseline+Spatial-DCT && 2.56 & 0.967 & 3.68 & 0.943 & 7.58 & 0.880 \\
Baseline+Spatial-DCT+Feature-DCT && \textbf{1.70} & \textbf{0.979} & \textbf{2.97} & \textbf{0.952} & \textbf{5.92} & \textbf{0.907} \\
\midrule
Baseline+Spatial-DCT+Feature-DCT & Click & 2.68 & 0.961 & 4.08 & 0.940 & 7.00 & 0.889 \\
\bottomrule
\end{tabular}
\end{table*}

\subsection{Implementation Details}
\label{ssec:details}
We formulate the training task as a binary segmentation problem and use binary cross-entropy loss for training.
We train all the models with a similar iterative training strategy in \cite{forte2020getting} on the 8498 images of SBD~\cite{hariharan2011sbd} and set the batch size to 8.
The input images are randomly resized from 0.75 to 1.25 of the original size and then randomly cropped at a fixed size of $256\times256$ pixels.
We further augment the training samples with horizontal flipping and color jitter.
We take ResNet50 pre-trained on ImageNet~\cite{imagenet} as backbone.
For optimization, we use Adam with $\beta_1=0.9, \beta_2=0.999$ and a learning rate of $10^{-4}$.
The learning rate is reduced by a factor of 0.1 every 10 epochs, and training completes after 20 epochs.

\begin{table}
\caption{Comparison of the mean number of clicks~(mNoC) on different datasets~\cite{rother2004grabcut,mcguinness2010berkeley,perazzi2016davis}.
* indicates the use of click and drag scheme to get the diffusion radius.}
\vspace{5pt}
\label{tab:comparison_of_noc}
\centering
\footnotesize 
\begin{tabular}{lccc}
\toprule
\multirow{2}{*}{Method} & GrabCut & Berkeley & DAVIS \\
& NoC @ 90\% & NoC @ 90\% & NoC @ 90\% \\
\midrule
Graph cut~\cite{boykov2001interactive} & 11.10 & 14.33 & 17.41 \\
% Growcut~\cite{vezhnevets2005growcut} & 16.74 & 18.25 & - \\
Random walker~\cite{grady2006random} & 12.30 & 14.02 & 18.31 \\
Geodesic matting~\cite{bai2007geodesic} & 12.44 & 15.96 & 19.50 \\
ESC~\cite{gulshan2010geodesic} & 9.20 & 12.11 & 17.70 \\
GSC~\cite{gulshan2010geodesic} & 9.12 & 12.57 & 17.52 \\
\midrule
DOS~\cite{xu2016deep} & 6.04 & 8.65 & 12.58 \\
RIS-Net~\cite{liew2017regional} & 5.00 & 6.03 & - \\
IIS-LD~\cite{li2018interactive} & 4.79 & - & 9.57 \\
CMG~\cite{majumder2019content} & 3.58 & 5.60 & - \\
BRS~\cite{jang2019interactive} & 3.60 & 5.08 & 8.24 \\
f-BRS~\cite{sofiiuk2020f} & 2.98 & 4.34 & 7.81 \\
FCA-Net~\cite{lin2020interactive} & 2.14 & 4.19 & 7.90 \\
\midrule
% DCT-Net  & 2.96 & 4.29 & 7.28 \\
DCT-Net & 2.68 & 4.08 & 7.00 \\
DCT-Net* & \textbf{1.70} & \textbf{2.97} & \textbf{5.92} \\
\bottomrule
\end{tabular}
\end{table}

\subsection{Results}
\label{ssec:Results}

\noindent
\textbf{Ablation study.} In Tab.~\ref{tab:ablation_arc}, we analyze the effectiveness of each component in our proposed method.
We take the basic segmentation network with Euclidean distance transform as our baseline model and then gradually equip the proposed components.
Overall, our proposed method is highly beneficial for the interactive segmentation model.

\vspace{3pt}
\noindent
\textbf{Comparison to the state-of-the-art.} We compare our results with existing State-of-the-Art methods on three standard benchmark datasets, GrabCut~\cite{rother2004grabcut}, Berkeley~\cite{mcguinness2010berkeley}, and DAVIS~\cite{perazzi2016davis}.
Tab.~\ref{tab:comparison_of_noc} shows the average number of clicks required to reach 90\% IoU threshold noted as NoC~@~90\%.
Our model requires 2.68 clicks and 4.08 clicks on GrabCut and Berkeley, respectively, when using click input only.
Under the Click and Drag scheme, it achieves the same threshold in only 1.98 clicks and 2.68 clicks, while the existing methods need more than 2 clicks and 4 clicks.
For DAVIS, we can reach 90\% IoU threshold with less than 7 clicks and achieve a relative improvement of 20\%.
Our method achieves the lowest number of clicks required to reach the IoU threshold for all datasets, whether using Auto-Drag-Head or Click-and-Drag scheme to determine diffusion distance dynamically.

\begin{figure}[t]
\centering
\setlength\tabcolsep{1pt}
\begin{tabular}{@{}ccccc@{}}
    1 click & 2 clicks & 3 clicks & 4 clicks & 5 clicks \\
    \includegraphics[width=0.18\linewidth]{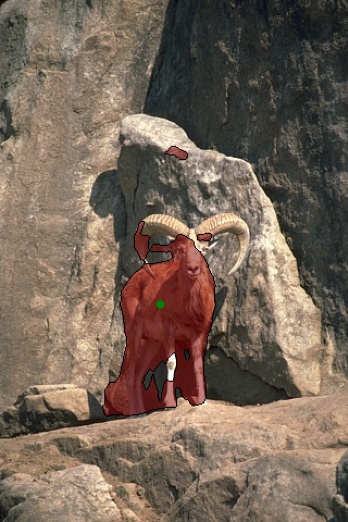} &
    \includegraphics[width=0.18\linewidth]{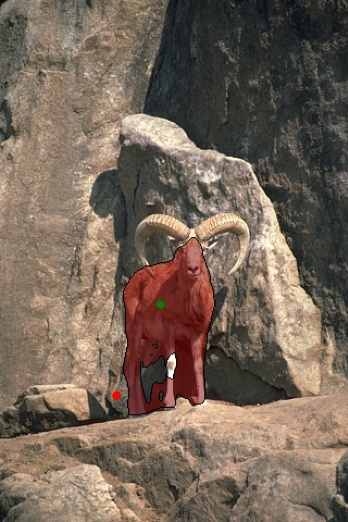} &
    \includegraphics[width=0.18\linewidth]{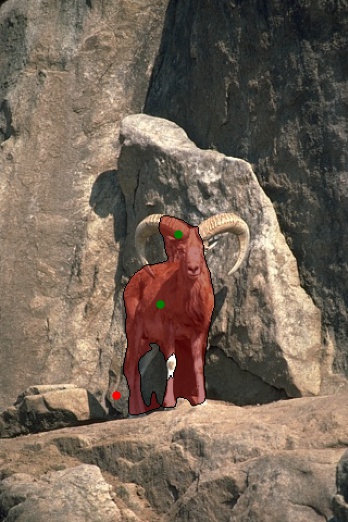} &
    \includegraphics[width=0.18\linewidth]{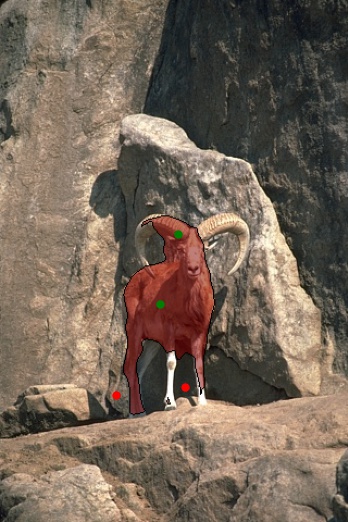} &
    \includegraphics[width=0.18\linewidth]{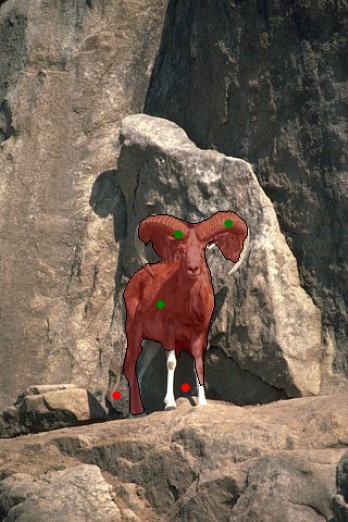} \\
    \includegraphics[width=0.18\linewidth]{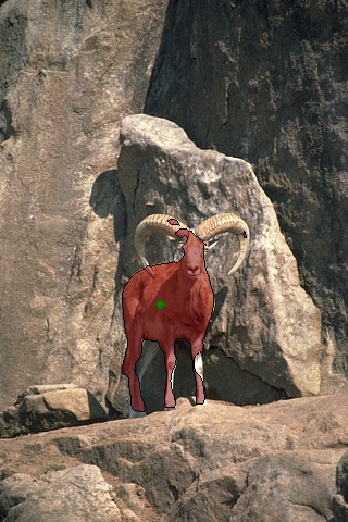} &
    \includegraphics[width=0.18\linewidth]{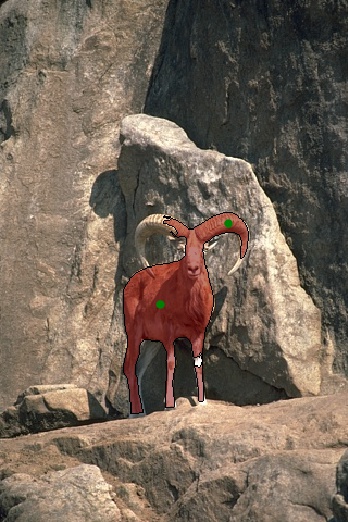} &
    \includegraphics[width=0.18\linewidth]{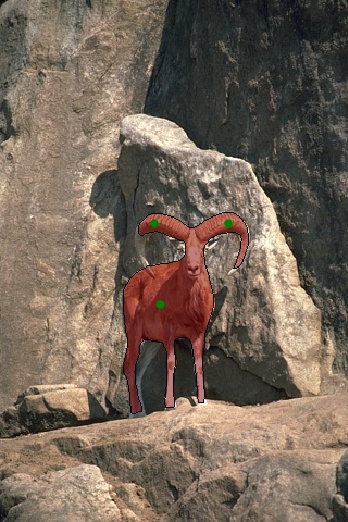} &
    \includegraphics[width=0.18\linewidth]{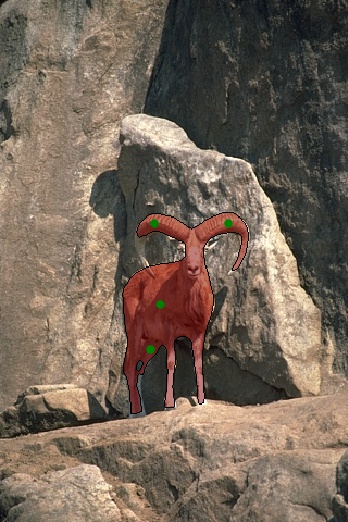} &
    \includegraphics[width=0.18\linewidth]{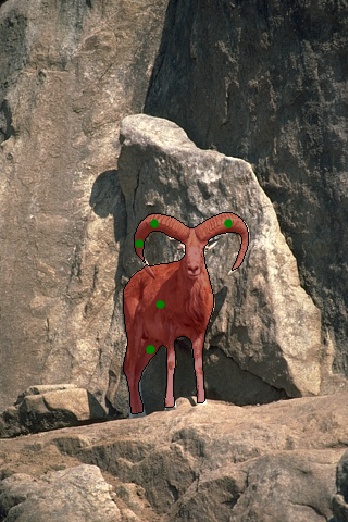} \\
\end{tabular}
\caption{Qualitative comparison between our baseline and full model for the first 5 clicks. Green points are positive clicks, red points are negative clicks, objects are overlaid with mask in dark red.}
\label{fig:quality_5}
\end{figure}

\section{Conclusion}
\label{sec:conclusion}
In this paper, we contribute to improving interactive object segmentation by a novel algorithm that reaches a good balance in human-machine collaboration.
Specifically, we propose the Dynamic Click Transform Network~(DCT-Net), which consists of a Spatial-DCT and a Feature-DCT to apply anisotropic diffusion for individual clicks and aggregate the corresponding feature to adjust the distribution of the original feature map in a forward pass, respectively.
%
%Spatial-DCT considers each click individually and applies anisotropic diffusion on the click maps.
%
%Practically, the diffusion radius can be obtained from Click and Drag scheme or Auto Drag Head, determining from the usage scenario.
%
%As for Feature-DCT, we aggregate the features corresponding to user clicks and adjust the distribution of the original feature map according to this feature in a forward pass.
%
The conducted experiments demonstrate the effectiveness of our proposed method and show the state-of-the-art performances over three standard interactive segmentation benchmarks.

\vspace{3pt}
\noindent
\textbf{Acknowledgement}
This research was supported in part by the Ministry of Science and Technology of Taiwan (MOST 110-2218-E-002-025-), National Taiwan University (NTU-108L104039), Intel Corporation, Delta Electronics and Compal Electronics.

\bibliographystyle{IEEEbib}
% \small
\bibliography{strings,refs}

\begin{thebibliography}{10}

\bibitem{rother2004grabcut}
Carsten Rother, Vladimir Kolmogorov, and Andrew Blake,
\newblock ``{"GrabCut"} interactive foreground extraction using iterated graph
  cuts,''
\newblock {\em ACM Transactions on Graphics (TOG)}, vol. 23, no. 3, pp.
  309--314, 2004.

\bibitem{xu2016deep}
Ning Xu, Brian Price, Scott Cohen, Jimei Yang, and Thomas~S Huang,
\newblock ``Deep interactive object selection,''
\newblock in {\em Proceedings of IEEE Conference on Computer Vision and Pattern
  Recognition (CVPR)}, 2016, pp. 373--381.

\bibitem{jang2019interactive}
Won-Dong Jang and Chang-Su Kim,
\newblock ``Interactive image segmentation via backpropagating refinement
  scheme,''
\newblock in {\em Proceedings of IEEE Conference on Computer Vision and Pattern
  Recognition (CVPR)}, 2019, pp. 5297--5306.

\bibitem{li2018interactive}
Zhuwen Li, Qifeng Chen, and Vladlen Koltun,
\newblock ``Interactive image segmentation with latent diversity,''
\newblock in {\em Proceedings of IEEE Conference on Computer Vision and Pattern
  Recognition (CVPR)}, 2018, pp. 577--585.

\bibitem{liew2017regional}
JunHao Liew, Yunchao Wei, Wei Xiong, Sim-Heng Ong, and Jiashi Feng,
\newblock ``Regional interactive image segmentation networks,''
\newblock in {\em Proceedings of IEEE International Conference on Computer
  Vision (ICCV)}. IEEE, 2017, pp. 2746--2754.

\bibitem{grady2006random}
Leo Grady,
\newblock ``Random walks for image segmentation,''
\newblock {\em IEEE Transactions on Pattern Analysis and Machine Intelligence
  (TPAMI)}, vol. 28, no. 11, pp. 1768--1783, 2006.

\bibitem{gulshan2010geodesic}
Varun Gulshan, Carsten Rother, Antonio Criminisi, Andrew Blake, and Andrew
  Zisserman,
\newblock ``Geodesic star convexity for interactive image segmentation,''
\newblock in {\em Proceedings of IEEE Conference on Computer Vision and Pattern
  Recognition (CVPR)}. IEEE, 2010, pp. 3129--3136.

\bibitem{boykov2001interactive}
Yuri~Y Boykov and M-P Jolly,
\newblock ``Interactive graph cuts for optimal boundary \& region segmentation
  of objects in nd images,''
\newblock in {\em Proceedings of IEEE International Conference on Computer
  Vision (ICCV)}. IEEE, 2001, vol.~1, pp. 105--112.

\bibitem{bai2007geodesic}
Xue Bai and Guillermo Sapiro,
\newblock ``A geodesic framework for fast interactive image and video
  segmentation and matting,''
\newblock in {\em Proceedings of IEEE International Conference on Computer
  Vision (ICCV)}. IEEE, 2007, pp. 1--8.

\bibitem{long2015fully}
Jonathan Long, Evan Shelhamer, and Trevor Darrell,
\newblock ``Fully convolutional networks for semantic segmentation,''
\newblock in {\em Proceedings of IEEE Conference on Computer Vision and Pattern
  Recognition (CVPR)}, 2015, pp. 3431--3440.

\bibitem{mahadevan2018iteratively}
Sabarinath Mahadevan, Paul Voigtlaender, and Bastian Leibe,
\newblock ``Iteratively trained interactive segmentation,''
\newblock in {\em Proceedings of British Machine Vision Conference (BMVC)},
  2018.

\bibitem{majumder2019content}
Soumajit Majumder and Angela Yao,
\newblock ``Content-aware multi-level guidance for interactive instance
  segmentation,''
\newblock in {\em Proceedings of IEEE Conference on Computer Vision and Pattern
  Recognition (CVPR)}, 2019, pp. 11602--11611.

\bibitem{sofiiuk2020f}
Konstantin Sofiiuk, Ilia Petrov, Olga Barinova, and Anton Konushin,
\newblock ``{f-BRS}: Rethinking backpropagating refinement for interactive
  segmentation,''
\newblock in {\em Proceedings of IEEE Conference on Computer Vision and Pattern
  Recognition (CVPR)}, 2020, pp. 8623--8632.

\bibitem{pont2015semi}
Jordi Pont-Tuset, Miquel~A Farr{\'e}, and Aljoscha Smolic,
\newblock ``Semi-automatic video object segmentation by advanced manipulation
  of segmentation hierarchies,''
\newblock in {\em 2015 13th International Workshop on Content-Based Multimedia
  Indexing (CBMI)}. IEEE, 2015, pp. 1--6.

\bibitem{ulyanov2016instance}
Dmitry Ulyanov, Andrea Vedaldi, and Victor Lempitsky,
\newblock ``Instance normalization: The missing ingredient for fast
  stylization,''
\newblock {\em arXiv preprint arXiv:1607.08022}, 2016.

\bibitem{forte2020getting}
Marco Forte, Brian Price, Scott Cohen, Ning Xu, and Fran{\c{c}}ois Piti{\'e},
\newblock ``Getting to 99\% accuracy in interactive segmentation,''
\newblock {\em arXiv preprint arXiv:2003.07932}, 2020.

\bibitem{mcguinness2010berkeley}
Kevin McGuinness and Noel~E O’connor,
\newblock ``A comparative evaluation of interactive segmentation algorithms,''
\newblock {\em Pattern Recognition}, vol. 43, no. 2, pp. 434--444, 2010.

\bibitem{perazzi2016davis}
Federico Perazzi, Jordi Pont-Tuset, Brian McWilliams, Luc Van~Gool, Markus
  Gross, and Alexander Sorkine-Hornung,
\newblock ``A benchmark dataset and evaluation methodology for video object
  segmentation,''
\newblock in {\em Proceedings of IEEE Conference on Computer Vision and Pattern
  Recognition (CVPR)}, 2016, pp. 724--732.

\bibitem{hariharan2011sbd}
Bharath Hariharan, Pablo Arbel{\'a}ez, Lubomir Bourdev, Subhransu Maji, and
  Jitendra Malik,
\newblock ``Semantic contours from inverse detectors,''
\newblock in {\em Proceedings of IEEE International Conference on Computer
  Vision (ICCV)}. IEEE, 2011, pp. 991--998.

\bibitem{imagenet}
J.~{Deng}, W.~{Dong}, R.~{Socher}, L.~{Li}, {Kai Li}, and {Li Fei-Fei},
\newblock ``Imagenet: A large-scale hierarchical image database,''
\newblock in {\em Proceedings of IEEE Conference on Computer Vision and Pattern
  Recognition (CVPR)}, 2009, pp. 248--255.

\bibitem{lin2020interactive}
Zheng Lin, Zhao Zhang, Lin-Zhuo Chen, Ming-Ming Cheng, and Shao-Ping Lu,
\newblock ``Interactive image segmentation with first click attention,''
\newblock in {\em Proceedings of IEEE Conference on Computer Vision and Pattern
  Recognition (CVPR)}, 2020, pp. 13339--13348.

\end{thebibliography}

\end{document}